\documentclass[runningheads]{llncs}

\usepackage{eccv}

\usepackage{eccvabbrv}

\usepackage{graphicx}
\usepackage{booktabs}

\usepackage[accsupp]{axessibility}  

\usepackage{pifont}

\usepackage{hyperref}

\usepackage{orcidlink}

\usepackage{siunitx}
\usepackage{texnames}
\usepackage{bm,bbm}
\usepackage{orcidlink}

\usepackage{booktabs}
\usepackage{graphicx}

\usepackage{tikz}
\usetikzlibrary{positioning}
\usetikzlibrary{calc}

\usepackage[percent]{overpic}

\usepackage{multirow}
\usepackage{adjustbox}
\usepackage[table]{xcolor}
\usepackage{tabularx}
\usepackage{makecell}

\newcommand{\highlighttext}[2][yellow]{%
  \begingroup
  \setlength{\fboxsep}{1.5pt}%
  \colorbox{#1}{#2}%
  \endgroup
}

\DeclareMathAlphabet{\mathcal}{OMS}{cmsy}{m}{n}

\begin{document}

\title{SeasonStereo: Robust Dense Stereo Matching for Multi-Date Satellite Imagery via Generative AI} 

\titlerunning{SeasonStereo}

\author{Álvaro Díaz-Laureano\inst{1}\and
Roger Marí\inst{1}\and\\
Elías Masquil\inst{2}\and
Pablo Arias\inst{3}\and
Gabriele Facciolo\inst{4,5}
}
\authorrunning{Á.~Díaz-Laureano et al.}

\institute{\textsuperscript{1 }Eurecat, Multimedia Technologies, Barcelona, Spain \\
\textsuperscript{2 }IIE, Facultad de Ingeniería, Universidad de la República, Uruguay \\
\textsuperscript{3 }Dept. of Engineering, Universitat Pompeu Fabra, Barcelona, Spain \\
\textsuperscript{4 }Université Paris-Saclay, CNRS, ENS Paris-Saclay, Centre Borelli, France \\
\textsuperscript{5} Institut Universitaire de France \\
\email{\{alvaro.diaz,roger.mari\}@eurecat.org}}

\maketitle

\begin{abstract}
Accurate 3D reconstruction from satellite imagery typically relies on near-simultaneous stereo pairs, limiting its applicability to diachronic settings where multi-date images exhibit varying seasonal and illumination conditions. Training dense stereo matching models robust to appearance changes is a long-standing challenge, as aligned multi-date imagery and ground-truth geometry are costly to obtain at scale. We propose SeasonStereo, a scalable framework that addresses disparity estimation from diachronic satellite images by training on synthetic image pairs with controlled seasonal appearance variation, while leveraging zero-shot geometric priors from foundation models. SeasonStereo matches the accuracy of state-of-the-art LiDAR-supervised models, while producing sharper geometric details without requiring aligned real multi-date training products or LiDAR-derived labels. As a result, SeasonStereo offers a practical path toward large-scale 3D reconstruction from heterogeneous satellite images with reduced supervision cost.


\end{abstract}

\definecolor{myyellow}{RGB}{255,217,102}
\definecolor{myblue}{RGB}{159,197,232}
\definecolor{mygreen}{RGB}{217,234,211}

\tikzset{
  figlabel/.style={font=\fontfamily{ptm}\selectfont\footnotesize}
}

\newcommand{\smalltilde}{\raisebox{0.15ex}{\tiny$\sim$}}
\begin{figure}[t]
\centering
\begin{tabular}{@{}c@{}}
 \begin{tikzpicture}[
  every node/.style={font=\rmfamily\normalsize}
 ]
  \node[anchor=south west, inner sep=0] (img) at (0,0){
  \includegraphics[
    draft=false,
    width=\linewidth
  ]{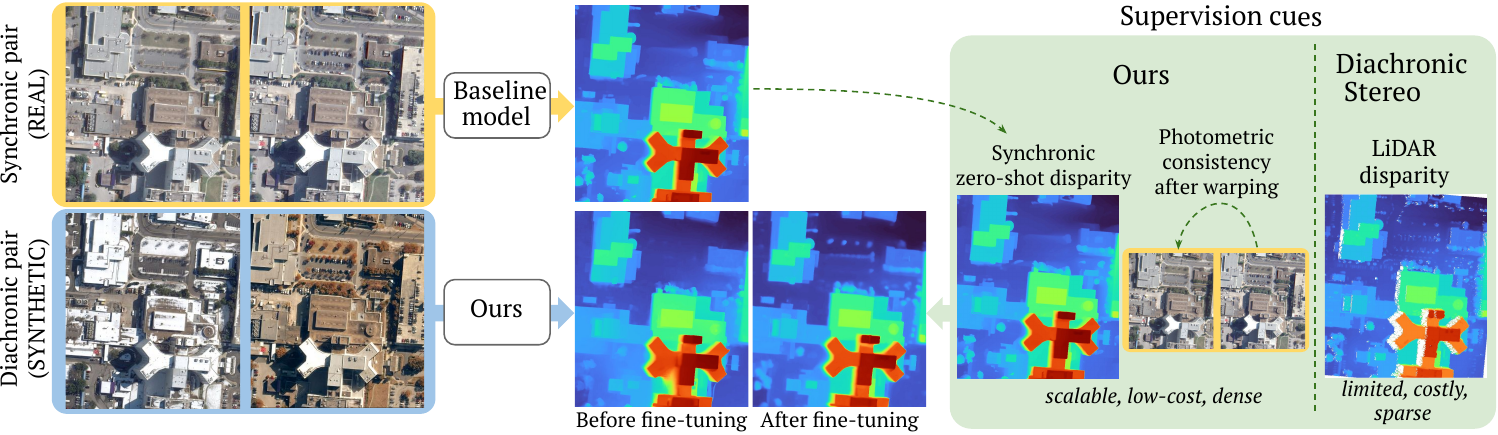}};
  \begin{scope}[x={(img.south east)},y={(img.north west)}]
    \node[text width=1cm] at (0.98, 0.78)
      {\tiny \cite{masquil2026diachronic}};
  \end{scope}
 \end{tikzpicture}
\end{tabular}
\caption{SeasonStereo learns to estimate disparity from multi-date satellite images by training on \highlighttext[myblue]{synthetic diachronic pairs} generated from \highlighttext[myyellow]{real synchronic pairs}. Supervision cues are derived from the corresponding synchronic pairs, combining zero-shot disparity estimates with photometric consistency after disparity-based warping. Unlike existing methods~\cite{masquil2026diachronic}, SeasonStereo avoids expensive LiDAR-based supervision.}
\label{fig:teaser}
\end{figure}

\section{Introduction}
\label{sec:intro}

Traditionally, 3D reconstruction from satellite images relies on stereo pairs acquired under near-simultaneous or \textit{synchronic} conditions. Synchronic images exhibit strong photometric consistency, allowing existing approaches for disparity estimation to achieve high levels of detail and accuracy~\cite{albanwan2022comparative, mari2022disparity}. However, performance degrades markedly when images deviate from such controlled conditions, highlighting a fundamental robustness gap.
Temporally distant images under \textit{diachronic} conditions, affected by strong seasonal changes and illumination differences, remain a relevant challenge, as photometric discrepancies hinder existing classical and learned approaches~\cite{masquil2026diachronic}. 

The development of robust dense stereo matching models for satellite imagery has long been constrained by data availability. High-resolution imagery is expensive, and acquiring reliable 3D ground-truth geometry such as LiDAR is even more resource-intensive. These requirements are difficult to satisfy across large geographic regions, limiting the deployment of these models at global scale. As a result, existing learned approaches~\cite{wu2021new, albanwan2022comparative, mari2022disparity, masquil2026diachronic} remain tied to specific locations or data conditions, underscoring the need for scalable and reproducible frameworks across diverse geographic and temporal settings.

In this work, we show that recent advances in image generation and foundation stereo models provide a scalable route to robust diachronic stereo matching in multi-date satellite imagery. Generative models can synthesize and manipulate images to simulate appearance variation~\cite{google_gemini3proimage}, while foundation stereo models provide strong zero-shot geometric cues for supervision~\cite{monster, monster++, stereo-anywhere, foundationstereo}.

We introduce SeasonStereo, a novel framework for disparity estimation from multi-date satellite images, trained on real image pairs and synthetic pairs with simulated seasonal variation. Synthetic samples, generated from real synchronic stereo pairs, augment the training data used to fine-tune a foundation stereo model for diachronic matching. As a result, SeasonStereo provides a practical and scalable route to large-scale 3D reconstruction from heterogeneous satellite images with substantially reduced supervision cost.
Our key contributions are:
\begin{itemize}
\item A framework for dense stereo matching of multi-date satellite images that uses synthetic seasonal appearance to remove the need for resource-intensive LiDAR-based geometric supervision or real diachronic products.
\item A multi-term training objective that combines zero-shot geometric priors from foundation models for global geometric accuracy with photometric and smoothness terms that enhance the sharpness of building contours.
\item A large-scale satellite stereo dataset of aligned real and synthetic image pairs in synchronic and diachronic settings with corresponding disparity maps. 
\end{itemize}
Fig.~\ref{fig:teaser} shows an overview of the method. We assess SeasonStereo on test sets of synchronic and diachronic WorldView-3 images of different geographic areas~\cite{masquil2026diachronic}.
Project page: \url{https://multimedia-eurecat.github.io/SeasonStereo}.
\section{Related Work}
\label{sec:relatedwork}

Dense stereo matching typically assumes photometric consistency between synchronic views. In satellite imagery, however, images are often acquired days or weeks apart, causing diachronic appearance variation. Despite its practical relevance, diachronic satellite stereo remains largely underexplored. To the best of our knowledge, Masquil et al.~\cite{masquil2026diachronic} provide the only model explicitly targeting diachronic stereo matching, using curated real pairs with LiDAR-based supervision. Our work follows this line of research, proposing a more scalable framework.

\subsection{Dense Stereo Matching}

Given a rectified image pair $(I_L, I_R)$, dense stereo matching estimates pixel-wise correspondences between views~\cite{scharstein2002classic}. Rectification constrains correspondences to horizontal displacements, or disparities. The left-view disparity map $D$ satisfies
\begin{equation}
I_L(x,y) \,\, {\Longleftrightarrow} \,\,  I_R(x-D(x,y), y),
\label{eq:disp}
\end{equation}
where $D(x,y)$ is the horizontal offset from pixel $(x,y)$ in $I_L$ to its match in $I_R$.

Traditional pipelines rely on handcrafted correspondence costs such as normalized cross-correlation or Census~\cite{census}. Semi-Global Matching (SGM)~\cite{sgm} became a dominant classical baseline by aggregating costs along multiple one-dimensional paths, achieving a strong accuracy-efficiency trade-off. SGM variants~\cite{facciolo2015mgm, dumas2022improving} remain standard in satellite stereo pipelines~\cite{beyer2018ames, franchis2014s2p}. However, handcrafted algorithms are sensitive to radiometric changes, low texture, occlusions, and depth discontinuities~\cite{tosi2025survey}. Early end-to-end deep learning methods unified learned cost volume construction, regularization and disparity regression~\cite{kendall2017end, laga2020survey}. Later models improved this paradigm through multi-scale context~\cite{zhao2017pyramid}, hierarchical cost volumes~\cite{yang2019hierarchical}, guided aggregation~\cite{zhang2019ganet}, or adaptive disparity search~\cite{duggal2019deeppruner}. In remote-sensing imagery, such models can outperform classical matching after target-domain adaptation but remain sensitive to domain shift~\cite{wu2021new,mari2022disparity}.

Recent learned stereo models increasingly combine recurrent matching with monocular cues. Building on RAFT-Stereo~\cite{raft-stereo}, which refines disparities through iterative recurrent updates, methods such as MonSter~\cite{monster}, Stereo Anywhere~\cite{stereo-anywhere}, FoundationStereo~\cite{foundationstereo}, and MonSter++~\cite{monster++} integrate monocular priors~\cite{depth-anything} to improve stereo matching robustness in   ambiguous regions. These methods mainly differ in how monocular and stereo cues are fused and refined, but share the goal of improving generalization beyond standard supervised stereo datasets. Among them, MonSter++ extends the MonSter family toward a unified foundation model for stereo and multi-view depth estimation.

In parallel, self-supervised training objectives have explored reducing the dependence on dense disparity or depth labels. Photometric reconstruction losses and left-right consistency, popularized by self-supervised monocular depth frameworks trained from stereo pairs such as Monodepth~\cite{monodepth}, were further improved in Monodepth2~\cite{monodepth2} through minimum reprojection and multi-scale training strategies. The choice of reconstruction penalty, including L1, SSIM~\cite{wang2004ssim}, or combinations thereof, also affects convergence and geometric sharpness. Such formulations are attractive for remote sensing because dense ground-truth geometry is costly to obtain, and they can complement supervised losses by encouraging image-consistent structure at object boundaries.

\subsection{3D Reconstruction from Satellite Imagery}

Satellite disparity maps are triangulated into Digital Surface Models (DSMs), typically within end-to-end pipelines handling image rectification, matching, and triangulation. Classical satellite stereo pipelines such as ASP~\cite{beyer2018ames}, MicMac~\cite{rupnik2017micmac}, CARS~\cite{michel2020cars}, S2P~\cite{franchis2014s2p} and variants~\cite{amadei2025s2p-hd, masquil2026deep-s2p} remain standard choices for synchronic imagery, but underlying matching algorithms are sensitive to diachronic inputs.

Because large-scale 3D annotations for training are costly to obtain in the satellite domain, several learning-based approaches avoid direct reconstruction from image pairs and focus on refining the outputs of classical stereo pipelines~\cite{stucker2022resdepth,bittner2018dsm,bittner2019late,bittner2018automatic}. ResDepth~\cite{stucker2022resdepth} learns residual corrections from coarse stereo DSM estimates and aligned imagery. Image evidence can help recover sharper man-made structures, such as building edges~\cite{bittner2019late,bittner2018automatic}. These approaches are therefore complementary to dense matching methods rather than replacements for them.

When multiple satellite acquisitions are available, reconstruction is commonly decomposed into a set of pairwise stereo problems followed by geometric fusion~\cite{facciolo2017automatic, qin2017automated, qin2022uncertainty, mari2021automatic}. Viewing angles, temporal proximity, and radiometric similarity are commonly used to prioritize pairs close to synchronic settings~\cite{facciolo2017automatic, gomez2023improving}. The resulting pairwise DSMs are spatially aligned and fused using robust aggregation strategies, such as median or confidence-based filtering~\cite{facciolo2017automatic,qin2019critical, kuschk2016spatially}. This multi-view stereo paradigm still depends heavily on finding sufficiently consistent image pairs, which limits its applicability to multi-date collections.

Recent self-supervised reconstruction methods based on Neural Radiance Fields~\cite{mari2022sat,mari2023multi} and 3D Gaussian Splatting~\cite{savant2025eogs,bournez2025eogs++} have extended satellite 3D reconstruction beyond explicit pairwise stereo by jointly optimizing appearance and geometry over multi-date image collections. These methods can produce highly accurate reconstructions, but they are computationally more demanding and typically require several observations of the same area. Sparse-view variants with depth-regularized formulations~\cite{zhang2023sparsesat} aim to improve reconstruction when fewer images are available; nevertheless, neural rendering approaches remain primarily designed for collection-level reconstruction, and are not a direct solution for geometry estimation from a single diachronic stereo pair.

\subsection{Real and Synthetic Data for Satellite Stereo}

Learning-based 3D reconstruction from satellite images has historically been constrained by limited public data~\cite{bosch2016}. This gap has been partly addressed by benchmarks such as the 2016 IARPA MVS Challenge~\cite{bosch2016},  DFC2019/US3D~\cite{bosch2019,lesaux2019} and CORE3D~\cite{brown2018large}, which provide high-resolution multi-view satellite imagery, predominantly from WorldView-3, together with LiDAR-derived geometric reference data. Several subsequent works have reprocessed these resources into stereo-oriented benchmarks such as SatStereo~\cite{patil2019new} and Masquil et al.~\cite{masquil2026diachronic}, which curate different splits of rectified image pairs under synchronic and diachronic appearance. More recent datasets, such as Stellar~\cite{patil2023stellar} and WHU-Stereo~\cite{li2023whu-stereo}, further expand geographic and sensor diversity. Nevertheless, these datasets remain based on real satellite acquisitions and external geometric supervision, and therefore inherit limitations related to ground-truth quality. In particular, LiDAR-derived supervision can contain voids, reduced density, rasterization artifacts near discontinuities, and temporal mismatches with the imagery due to scene changes between the acquisition dates~\cite{masquil2026diachronic,patil2019new}. These errors mainly affect high-frequency structures, biasing learned models toward smoother disparities and weaker building-boundary reconstruction~\cite{stucker2022resdepth,bittner2018dsm,bittner2018automatic}.

To overcome the scarcity of high-quality geometric supervision, recent work has explored synthetic alternatives such as rendered virtual scenes and generative models. Rendering-based datasets such as SkyScenes~\cite{khose2024skyscenes} for aerial scene understanding, and SatUnreal~\cite{kim2026satunreal}, which simulates satellite stereo acquisition geometry in Unreal Engine, show the potential of sim-to-real transfer. On the generative side, general-purpose image generators~\cite{betker2023dalle3,google_gemini3proimage} and satellite-domain-specific models~\cite{khanna2024diffusionsat,mari2025difussion,benidir2025change,jakubik2026terramind} offer a flexible path for scalable image synthesis and editing, as they are not restricted to explicitly modeled virtual scenes. However, their use for dense stereo matching requires generative control over spatial structure and multi-view consistency, which remains a challenge~\cite{wang2024diffusion}.

\section{Method}
\label{sec:method}
Given a pair of satellite images $I_L$ and $I_R$ observing the same area, SeasonStereo aims to estimate the corresponding disparity map $D$. While primarily targeting diachronic pairs, the method remains compatible with synchronic imagery, providing a unified framework for multi-date satellite stereo.

We describe SeasonStereo in three stages. First, we curate a strictly synchronic training set of real satellite stereo pairs, which serves as the basis for subsequent synthetic augmentation. Second, we use generative models to synthesize seasonal appearance variations, producing diachronic image pairs that remain geometrically aligned with their real synchronic counterparts. Third, we fine-tune a foundation stereo model with a multi-term supervision loss that leverages scalable training cues. 

\subsection{Curation of Synchronic Satellite Stereo Pairs via Similarity}
\label{subsec:dataset_curation}

The DFC2019 \textit{Track 3} dataset~\cite{bosch2019, lesaux2019} provides multi-date WorldView-3 image collections and LiDAR acquisitions over the Jacksonville (JAX) and Omaha (OMA) areas of interest (AOIs). Building on this dataset, Masquil et al.~\cite{masquil2026diachronic} released a stereo matching benchmark comprising rectified synchronic and diachronic image pairs, together with the associated rectification homographies. We use these resources to curate the necessary data for training SeasonStereo.

We focus on the \textit{synchronic-only} split from~\cite{masquil2026diachronic}, which contains 1,565 candidate image pairs. Although this split is designed to contain near-simultaneous acquisitions, visual inspection revealed that some pairs still exhibit noticeable photometric discrepancies. Since SeasonStereo relies on fully consistent real stereo pairs to derive reliable supervision cues, we apply a conservative multi-metric consensus strategy to identify a highly consistent subset for training.

\begin{figure}[t]
    \centering
    \includegraphics[width=1\linewidth]{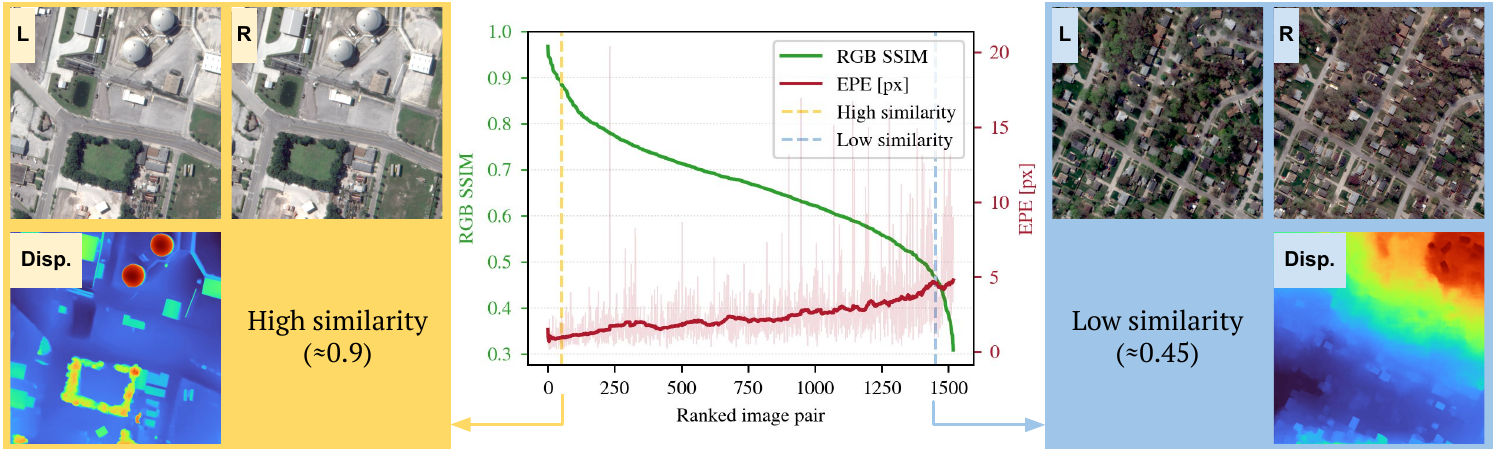}
    
    \vspace{-0.5em}
    
    \caption{Image pairs ranked by decreasing RGB-SSIM similarity~\eqref{eq:combined_score} (green), with the corresponding MonSter++ End-Point Error (EPE) against LiDAR-derived disparities (red). Blue and yellow lines correspond to low- and high-similarity pairs shown on either side of the plot. MonSter++ disparity accuracy degrades as similarity decreases.}
    \label{fig:similarity_epe}
\end{figure}

For each candidate pair $(I_L, I_R)$, we first estimate a disparity map $D$ using the pretrained foundation model MonSter++~\cite{monster++}. The disparity map is used to warp the right image onto the left image according to~\eqref{eq:disp}, yielding a reconstructed left image $\hat{I}_L$. We then compute photometric similarity between $I_L$ and $\hat{I}_L$ as a proxy for the consistency of the stereo pair. MonSter++ estimates are preferred over LiDAR-derived disparities for this warping step because LiDAR and image evidence may be misaligned due to non-coincident acquisition dates~\cite{masquil2026diachronic}. In addition, qualitative inspection showed that MonSter++ predictions produce sharper object boundaries and fewer warping artifacts.

Photometric consistency is evaluated using complementary low-level and learned measures.
First, we compute RGB 
distances between $I_L$ and $\hat{I}_L$ over valid warped pixels and convert them into bounded
similarity scores: $
S_{\mathrm{color}} = 1 -
\frac{\min(d_{\mathrm{color}}, d_{\max})}{d_{\max}},
$
where $d_{\mathrm{color}}$ is the mean of per-pixel distances for a given pair, and $d_{\max}$ is set to the 95th percentile of
$d_{\mathrm{color}}$ over all evaluated pairs. This color-based score is combined with SSIM~\cite{wang2004ssim}, normalized to $[0, 1]$ and denoted $S_{\mathrm{SSIM}}$, to obtain a single photometric consistency score:
\begin{equation}
    S_{\mathrm{comb}} = 0.5 (S_{\mathrm{color}}  + S_{\mathrm{SSIM}}).
    \label{eq:combined_score}
\end{equation}
In addition to this low-level score \eqref{eq:combined_score}, we compute a learned appearance similarity score $S_{\mathrm{DINOv3}}$ as the cosine
similarity between DINOv3 image features~\cite{simeoni2025dinov3}.
 
Candidate pairs are ranked independently by $S_{\mathrm{comb}}$ and $S_{\mathrm{DINOv3}}$. As shown in Fig.~\ref{fig:similarity_epe}, lower similarity scores generally correspond to higher MonSter++ disparity errors against LiDAR-derived reference disparities, highlighting the importance of selecting highly similar, synchronic pairs. To obtain the final subset of real synchronic pairs for SeasonStereo, we take the intersection of the 1,000 highest-scoring pairs under each similarity metric.

Overall, this data curation strategy yields a final subset of 871 highly consistent synchronic stereo pairs from the original 1,565 pairs in the \textit{synchronic-only} split of Masquil et al.~\cite{masquil2026diachronic}.

\definecolor{mygray}{gray}{0.25}
\begin{figure}[t]
    \centering
    \begin{adjustbox}{max width=\linewidth}
        \begin{tabular}{ccccc}
             \multicolumn{5}{c}{%
                \parbox{\linewidth}{%
                    \scriptsize \textcolor{mygray}{\textbf{Winter prompt:}  \textit{Satellite \textbf{winter} transform.
                    Edit only seasonal appearance.
                    Preserve exact camera viewpoint, framing, scale, alignment, and scene geometry.
                    Keep all buildings, roads, field boundaries, landmarks, river banks, shorelines,
                    and water body contours in exactly the same positions, widths, and shapes.
                    \textbf{Apply winter conditions only: snow on rooftops, fields,
                    and open ground; dormant or leafless vegetation;
                    lower sun angle shadows; (...)}}}
                }%
             } \\[4pt]
             \midrule 
             \includegraphics[width=0.2\linewidth]{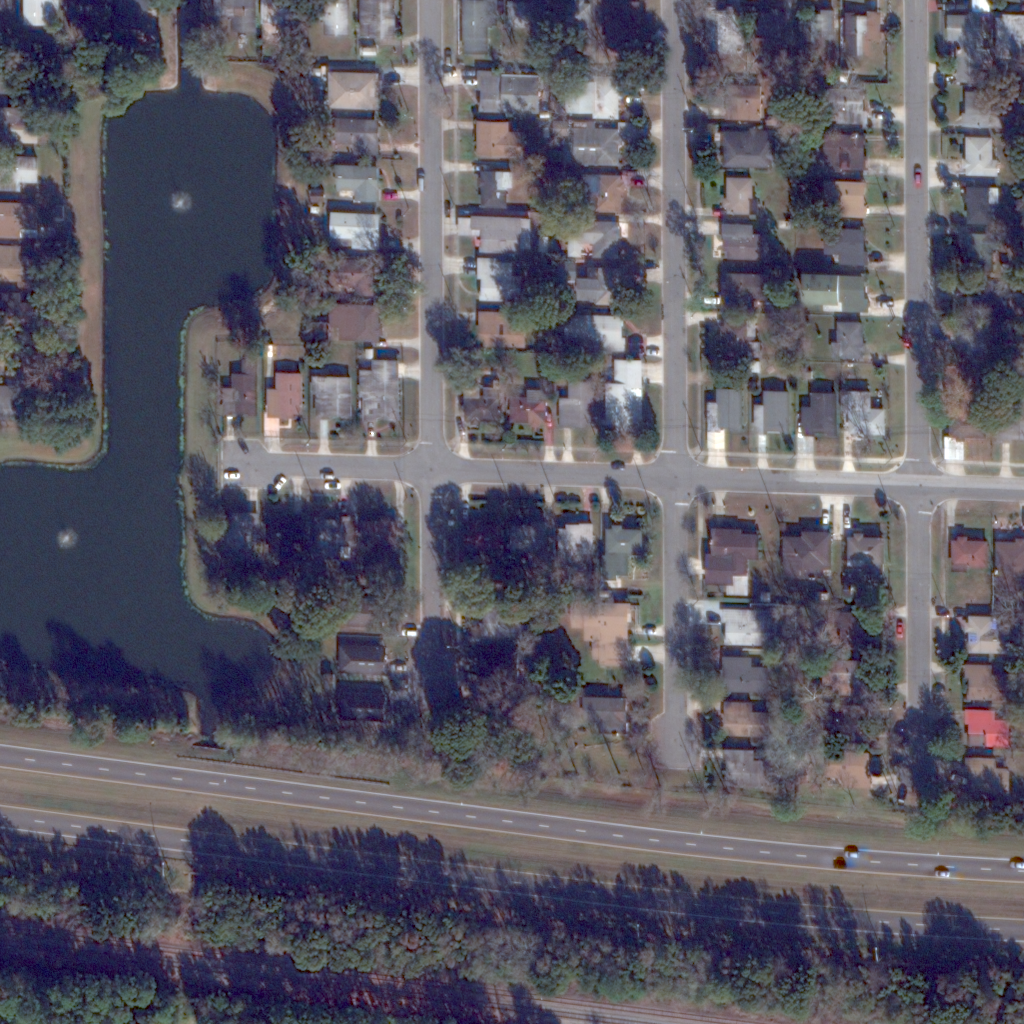}
             & \includegraphics[width=0.2\linewidth]{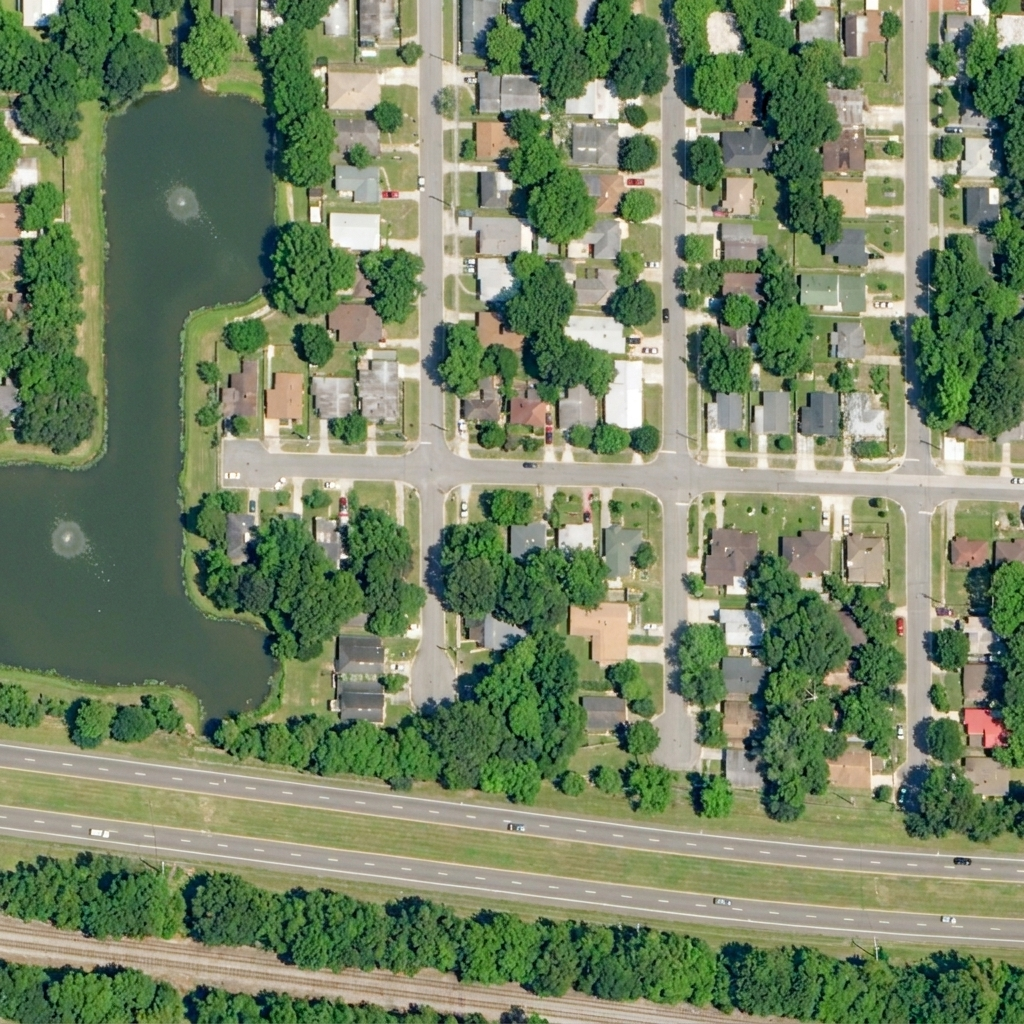}
             & \includegraphics[width=0.2\linewidth]{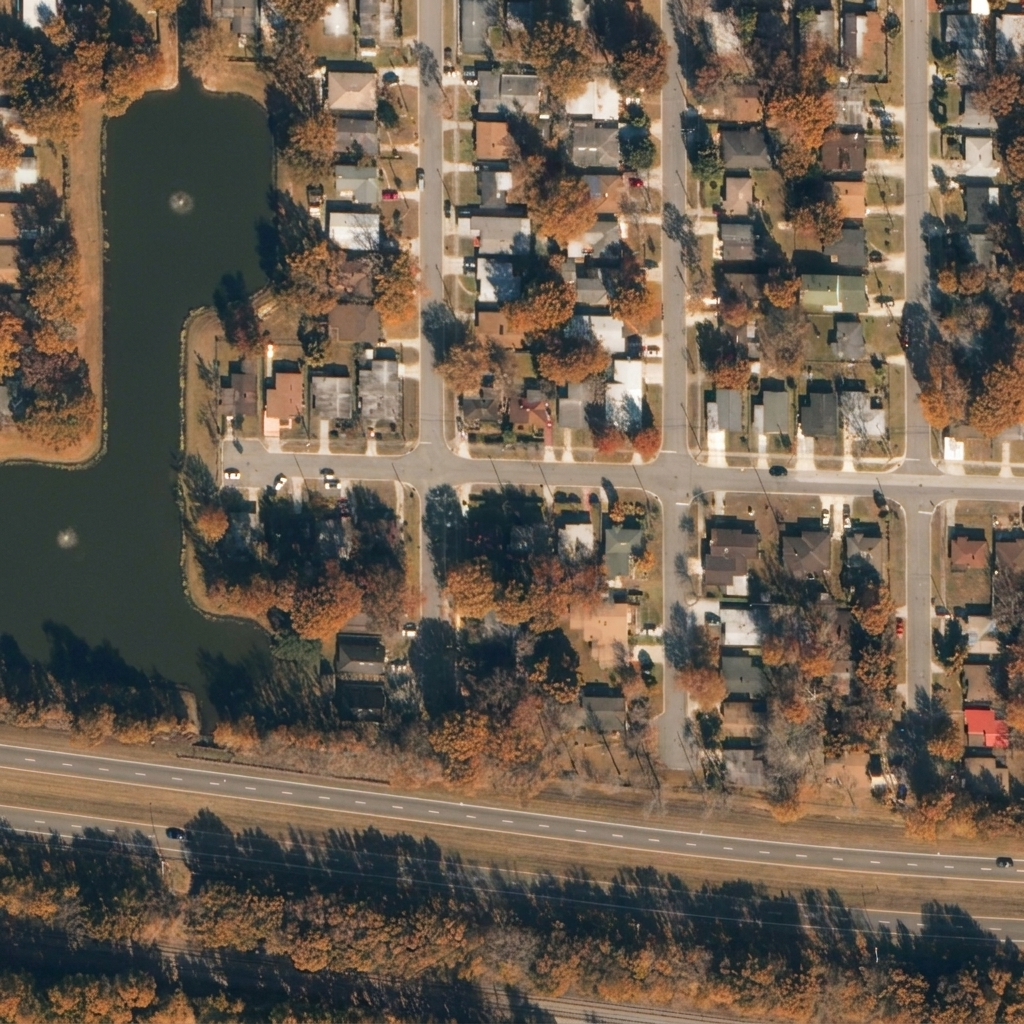}
             & \includegraphics[width=0.2\linewidth]{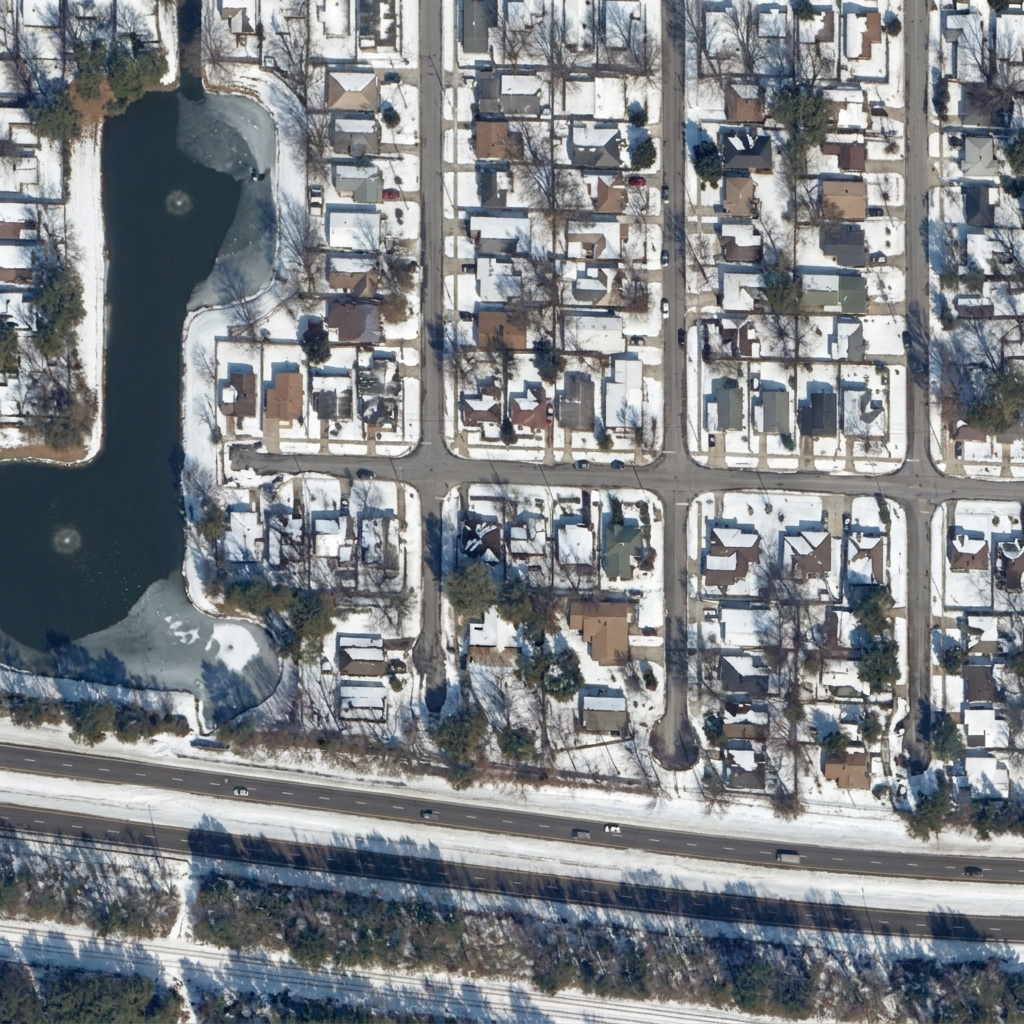}
             & \includegraphics[width=0.2\linewidth]{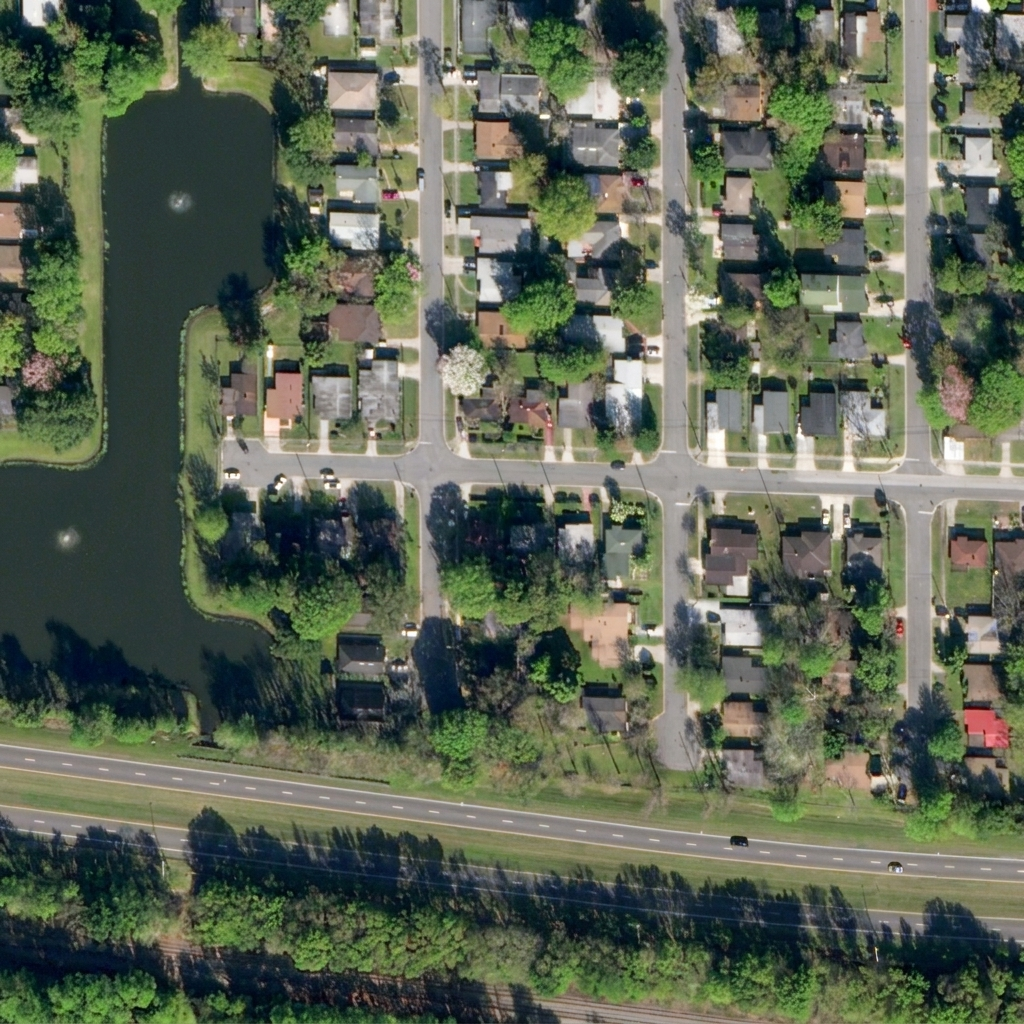} \\
             {\scriptsize Real image} & {\scriptsize Synthetic, summer} & {\scriptsize Synthetic, autumn} & {\scriptsize Synthetic, winter} & {\scriptsize Synthetic, spring}
        \end{tabular}
    \end{adjustbox}
    \vspace{-0.5em}
    \caption{Real image vs.\ corresponding geometry-preserving, aligned, seasonally diverse samples generated by Nano Banana Pro~\cite{google_gemini3proimage}. The prompt above was used for winter: bold text marks season-specific instructions; regular text is shared across all seasons.}
    \label{fig:synthetic_seasonal_examples}
\end{figure}

\begin{figure}[t]
\centering
\includegraphics[width=1\linewidth]{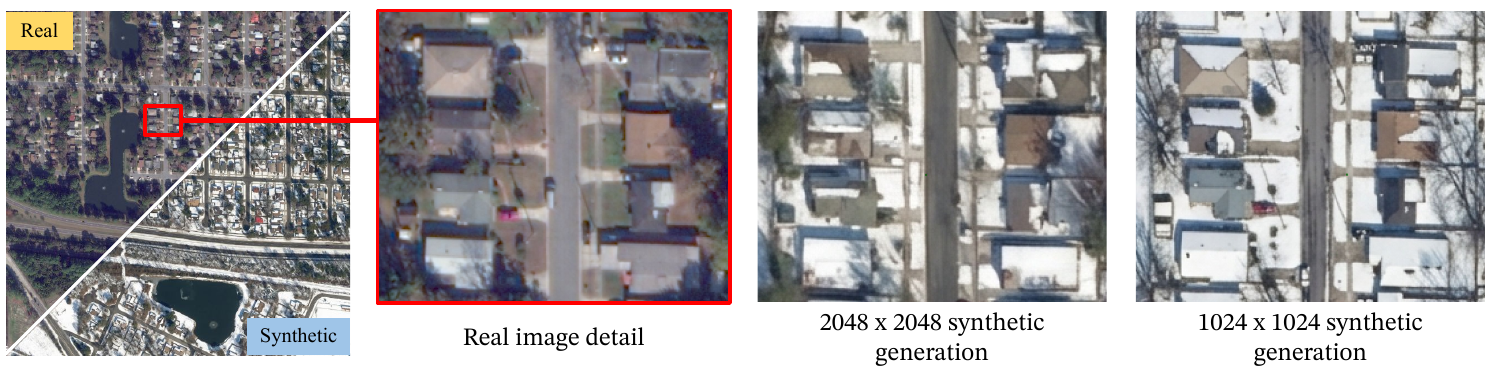}
\vspace{-0.5em}
\caption{Effect of input spatial extent on Nano Banana Pro generative structural fidelity. Generation from the full $2048{\times}2048$ image introduces local hallucinations and distorted building contours, while generation from the $1024{\times}1024$ center crops yields stronger geometric preservation.}
\label{fig:genai_comparison_resolution}
\end{figure}

\subsection{Geometry-Preserving Seasonal Synthesis of Satellite Images}
\label{subsec:seasonal_synthesis}

For each image in the 871 real synchronic pairs selected in Sec.~\ref{subsec:dataset_curation}, we generate four synthetic seasonally diverse variants using a text-conditioned generative model, as shown in Fig.~\ref{fig:synthetic_seasonal_examples}.  Together with the original images, this yields five appearance variants per view and 25 left–right combinations per pair (5 synchronic + 20 diachronic), expanding the training set to 21,775 pairs.

We use Nano Banana Pro~\cite{google_gemini3proimage} to generate geometry-preserving, seasonally diverse synthetic images aligned with the real images, as illustrated in Fig.~\ref{fig:synthetic_seasonal_examples}. We found the input spatial extent to be critical for preserving structural details: generation on the full $2048{\times}2048$ DFC2019 \textit{Track 3} images led to less faithful fine-scale geometry, whereas central $1024{\times}1024$ crops better preserved structures such as building contours, as shown in Fig.~\ref{fig:genai_comparison_resolution}.

We observe that water bodies and trees frequently produce geometrically inconsistent generations, making the inherited supervision cues unreliable in these regions. We therefore use an OpenEarthMap-pretrained SegFormer model~\cite{odil111_segformer_openearthmap_2024, xia2023openearthmap, xie2021segformer} to extract water and tree segmentation masks for each image. The corresponding pixels are marked as invalid and excluded from the loss computation during training (Fig.~\ref{fig:validity_mask}).

Finally, the rectification homographies provided by Masquil et al.~\cite{masquil2026diachronic} are used to rectify the synthetic samples and align them with the corresponding real synchronic pairs selected in Sec.~\ref{subsec:dataset_curation}. We release the resulting dataset of 21,775 rectified left–right image pairs, formed from real and synthetic seasonal variants across 108 JAX and OMA areas, together with the associated masks, MonSter++ zero-shot disparities, and LiDAR-derived disparities for the corresponding real pairs.

\subsection{Scalable Supervision for Diachronic Satellite Stereo Adaptation}
\label{subsec:loss_function}

MonSter++~\cite{monster++} is adopted as the disparity estimation backbone in SeasonStereo. It extends the MonSter architecture used in prior work on diachronic satellite stereo matching~\cite{masquil2026diachronic}.
MonSter++ follows RAFT-Stereo's sequence loss~\cite{raft-stereo} for training, supervising intermediate disparity estimates with increasing weights:
\begin{equation}
\mathcal{L} = \mathcal{L}^{(0)} + \sum_{k=1}^{N} w^{(k)} \mathcal{L}^{(k)},
\label{eq:sequence_loss}
\end{equation}
where $\mathcal{L}^{(k)}$ denotes the L1 disparity error between the prediction at iteration $k$ and the ground-truth disparity over valid pixels.
The weights follow an exponential schedule that up-weights later, more refined predictions, with $w^{(k)} = \gamma^{N-k}$ and $\gamma = 0.9^{15/(N-1)}$.

For SeasonStereo, we retain the sequence-loss formulation in~\eqref{eq:sequence_loss}, but replace the supervised per-iteration loss $\mathcal{L}^{(k)}$ with a multi-term objective:
\begin{equation}
\mathcal{L}^{(k)} = \lambda_{\mathrm{disp}}\,\mathcal{L}_{\mathrm{disp}}(\hat{D}^{(k)})
+ \lambda_{\mathrm{photo}}\,\mathcal{L}_{\mathrm{photo}}(\hat{D}^{(k)})
+ \lambda_{\mathrm{smooth}}\,\mathcal{L}_{\mathrm{smooth}}(\hat{D}^{(k)}),
\label{eq:seasonstereo_loss}
\end{equation}
where $\lambda_i$ are scalar weights. Following~\eqref{eq:seasonstereo_loss}, we fine-tune MonSter++ on the synchronic and diachronic stereo pairs derived in Sec.~\ref{subsec:seasonal_synthesis}. Crucially, \eqref{eq:seasonstereo_loss} relies on supervision cues automatically constructed from the real synchronic pairs, removing the need for resource-intensive geometric cues such as LiDAR-derived disparities used in~\cite{masquil2026diachronic}.

\subsubsection{Disparity Loss.}
\begin{figure}[t]
    \centering
    \includegraphics[width=1\linewidth, trim=0 0.3cm 0 0, clip]{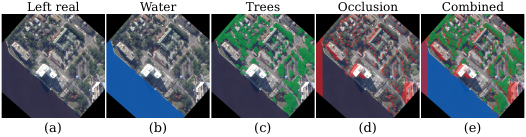}
    
    \vspace{-0.5em}
    
    \caption{Construction of left-view valid-pixel mask $\mathcal{V}$. Invalid regions comprise pixels outside the rectified image crop (black), water (blue), trees (green), and occluded or non-jointly-visible regions estimated from disparity (red). $\mathcal{V}$ excludes invalid pixels during training and evaluation.}
    \label{fig:validity_mask}
\end{figure}

SeasonStereo uses an L1 disparity loss as the main geometric supervision, without using LiDAR-derived disparities as dense training targets. A frozen pretrained MonSter++ model acts as a teacher, producing zero-shot disparities $\tilde{D}$ from the selected real synchronic pairs. Because the synthetic seasonal variants preserve geometry and alignment, these predictions can be transferred to supervise the corresponding synthetic diachronic pairs. The fine-tuned model predictions $\hat{D}^{(k)}$ are trained to match the teacher disparity $\tilde{D}$:
\begin{equation}
\mathcal{L}_{\mathrm{disp}}(\hat{D}^{(k)}) =
\frac{1}{|\mathcal{V}|}
\sum_{\mathbf{x} \in \mathcal{V}}
\left| \hat{D}^{(k)}(\mathbf{x}) - \tilde{D}(\mathbf{x}) \right|,
\label{eq:our_disp_loss}
\end{equation}
where $\mathbf{x}=(x,y)$ is the pixel location and $\mathcal{V}$ is the set of valid pixels (Fig.~\ref{fig:validity_mask}).

\subsubsection{Photometric Loss.}
Prior work has shown that image evidence can be used to refine fine-grained details of aligned geometry~\cite{stucker2022resdepth, bittner2019late, mari2023multi}. We include a photometric reconstruction loss that penalizes the error obtained by warping the right image into the left viewpoint using the predicted disparity. Directly applying this loss to diachronic pairs is ill-posed: seasonal appearance differences would persist after warping regardless of disparity accuracy. SeasonStereo avoids this issue by exploiting the fact that synthetic diachronic pairs share the geometry of the underlying real synchronic pair. Therefore, regardless of whether the input pair is synchronic or diachronic, the photometric loss is evaluated on the corresponding real synchronic pair:
\begin{equation}
\mathcal{L}_{\mathrm{photo}}(\hat{D}^{(k)}) =
\frac{1}{|\mathcal{V} \cap  \mathcal{V}_{\text{B}}|}
\sum_{\mathbf{x} \in \mathcal{V} \cap \mathcal{V}_{\text{B}}}
\ell_{\mathrm{photo}}
\left(
I^{\mathrm{real}}_L(\mathbf{x}),
\mathrm{warp}(I^{\mathrm{real}}_R, \hat{D}^{(k)})(\mathbf{x})
\right).
\label{eq:our_photo_loss}
\end{equation}
This allows the model to learn from diachronic inputs while the reconstruction signal is computed on appearance-consistent imagery. We adopt the per-pixel photometric error $\ell_{\mathrm{photo}}$ from the self-supervised depth estimation Monodepth framework~\cite{monodepth, monodepth2}, combining SSIM~\cite{wang2004ssim} and L1 differences:
\begin{equation}
\ell_{\mathrm{photo}}(I^{\mathrm{real}}_L, \hat{I}^{\mathrm{real}}_L) =
\alpha\,\frac{1 - \mathrm{SSIM}(I^{\mathrm{real}}_L, \hat{I}^{\mathrm{real}}_L)}{2} + (1-\alpha)\,|I^{\mathrm{real}}_L - \hat{I}^{\mathrm{real}}_L|_1, \quad \alpha = 0.85,
\end{equation}
where $I^{\mathrm{real}}_L$ and $\hat{I}^{\mathrm{real}}_L$ are the real left image and the reconstructed left image obtained after warping, i.e., $\hat{I}^{\mathrm{real}}_L = \mathrm{warp}(I^{\mathrm{real}}_R, \hat{D}^{(k)})$ in Eq.~\eqref{eq:our_photo_loss}.

\begin{figure}[t]
    \centering
    \includegraphics[width=1\linewidth]{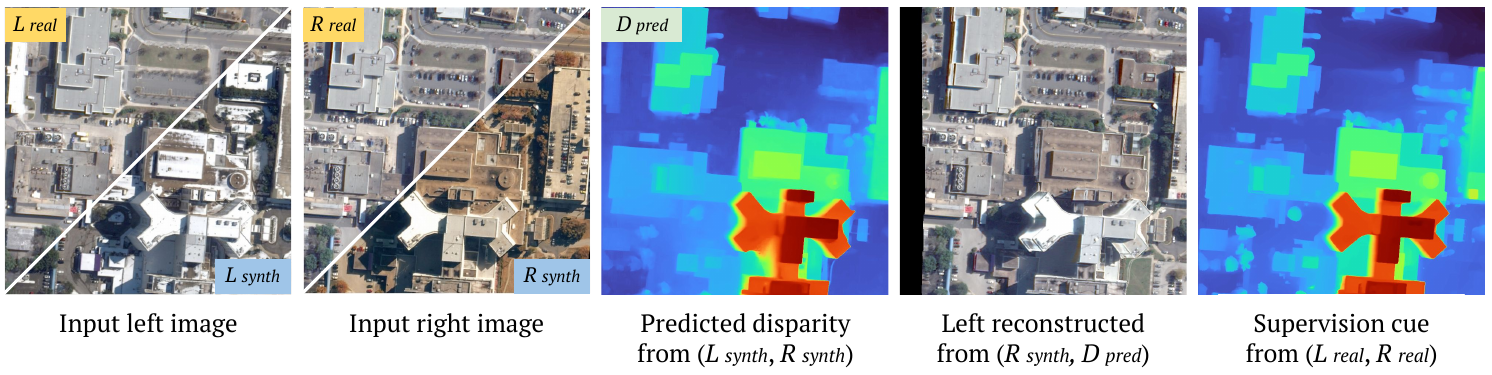}

    \vspace{-0.5em}

    \caption{SeasonStereo input images, predicted disparity on the diachronic pair before fine-tuning, teacher disparity derived from the real synchronic pair, and left-view reconstruction after warping the right view onto the left based on the predicted disparity.}
    \label{fig:loss_terms}
\end{figure}

Although the real synchronic pairs used for supervision are strongly consistent, they may still be acquired hours or days apart. Dynamic objects such as vehicles can therefore introduce noise in the photometric term $\mathcal{L}_{\mathrm{photo}}$. To mitigate this, we restrict the photometric loss to building regions, promoting sharper disparity estimates on permanent structures. Building masks, denoted $\mathcal{V}_{\text{B}}$ in Eq.~\eqref{eq:our_photo_loss}, are computed via OpenEarthMap-pretrained SegFormer~\cite{odil111_segformer_openearthmap_2024}, analogously to the water and tree masks discussed in Sec.~\ref{subsec:seasonal_synthesis}. Fig.~\ref{fig:loss_terms} illustrates the teacher disparity used for supervision in Eq.~\eqref{eq:our_disp_loss}, together with the left-view reconstruction used to compute the photometric loss in Eq.~\eqref{eq:our_photo_loss}.

\subsubsection{Smoothness Loss.}
Following Monodepth~\cite{monodepth}, we also use an auxiliary edge-aware smoothness regularizer to discourage spurious disparity fluctuations while allowing discontinuities at likely object boundaries. Disparity gradients are weighted by the inverse gradients of the left input image $I_L$:
\begin{equation}
\mathcal{L}_{\mathrm{smooth}}(\hat{D}^{(k)}) =
\frac{1}{|\mathcal{V}|}
\sum_{\mathbf{x} \in \mathcal{V}}
\left(
\left|\partial_x \hat{D}^{(k)}(\mathbf{x})\right| e^{-|\partial_x I_L(\mathbf{x})|}
+
\left|\partial_y \hat{D}^{(k)}(\mathbf{x})\right| e^{-|\partial_y I_L(\mathbf{x})|}
\right),
\label{eq:our_reg_loss}
\end{equation}
where large gradients reduce the penalty, preserving  disparity changes at image edges.

\subsubsection{Valid Pixel Mask.}
The loss terms in Eq.~\eqref{eq:our_disp_loss} and Eq.~\eqref{eq:our_reg_loss} use a valid-pixel mask $\mathcal{V}$. Eq.~\eqref{eq:our_photo_loss} is further restricted to building regions delineated by the intersection of $\mathcal{V}$ and the building mask $\mathcal{V}_B$. The valid-pixel mask $\mathcal{V}$ is obtained by excluding the unreliable regions shown in Fig.~\ref{fig:validity_mask}: black border areas falling outside the image crops after rectification; water and tree regions where synthetic generations may be geometrically inconsistent; and occluded pixels.

Occluded or non-jointly-visible pixels are  detected with a z‑buffer test based on the predicted disparity $\hat{D}$. Each pixel in the left image is shifted horizontally by its disparity; if it lands outside the right image, it is discarded. When several pixels map to the same location, we keep only the one with the largest disparity (the closest surface) and mark all others as occluded.
Fig.~\ref{fig:validity_mask} shows the resulting occlusion mask in red, mainly on building facades and near the image borders.

\section{Experiments}
\label{sec:experiments}

\label{subsec:data_eval_metrics}

Following Masquil et al.~\cite{masquil2026diachronic}, we evaluate on four test splits: \textbf{Jacksonville} (DFC2019~\cite{bosch2019}, predominantly synchronic image pairs), \textbf{Buenos Aires} (IARPA 2016 Multi-View Stereo Challenge~\cite{bosch2016}, soft diachronic pairs), and two \textbf{Omaha} sets (DFC2019~\cite{bosch2019}) containing synchronic and diachronic pairs, respectively.

Model performance is evaluated on the test sets using the altitude mean absolute error (MAE) of reconstructed DSMs from the predicted disparity maps.
All methods are evaluated using the validity mask $\mathcal{V}$ defined in Sec.~\ref{subsec:loss_function}. In particular, water and tree regions are excluded to avoid penalizing discrepancies caused by temporal differences between the image acquisitions and the LiDAR reference data.

We further conduct ablation studies to assess the impact of each loss term in SeasonStereo~\eqref{eq:seasonstereo_loss} and the contribution of synthetic diachronic training data. To isolate the latter, we train the full model based on~\eqref{eq:seasonstereo_loss} on three data subsets. Subset 1 (S1) contains only the curated real synchronic pairs from Sec.~\ref{subsec:dataset_curation}. Subset 2 (S2) augments S1 with generated synchronic synthetic pairs. Subset 3 (S3) augments S1 with both synchronic and diachronic synthetic pairs in equal proportion, while keeping the total augmentation budget identical to S2. All training runs use the full generated dataset described in Sec.~\ref{sec:method}, except those in the ablation study on the S1/S2/S3 subsets.

\subsection{Implementation Details} 
SeasonStereo is initialized from the mixed-domain MonSter++~\cite{monster++} checkpoint. We freeze the monocular depth prior branch and fine-tune only the stereo matching backbone and iterative update modules. The disparity, photometric, and smoothness loss weights in~\eqref{eq:seasonstereo_loss} are set to $\lambda_{\mathrm{disp}}=0.05$, $\lambda_{\mathrm{photo}}=0.1$ and $\lambda_{\mathrm{smooth}}=0.1$ across all experiments. We train for 50{,}000 iterations using the AdamW optimizer, a learning rate of $5\times10^{-4}$, a weight decay of $1\times10^{-5}$, a one-cycle schedule, and a batch size of 4 on an NVIDIA L40S GPU.
Data augmentation follows the RAFT-Stereo~\cite{raft-stereo} protocol adopted by MonSter~\cite{monster}, including cropping, color jittering, and horizontal flipping, with further details in~\cite{masquil2026diachronic}.

\newcommand{\expGT}{$\mathcal{L}_{\mathrm{LiDAR}}$}
\newcommand{\expPseudoGT}{$\mathcal{L}_{\mathrm{disp}}$}
\newcommand{\expPseudoGTPhoto}{$\mathcal{L}_{\mathrm{disp}} + \mathcal{L}_{\mathrm{photo}}$}
\newcommand{\expPseudoGTPhotoSmooth}{$\mathcal{L}_{\mathrm{disp}} + \mathcal{L}_{\mathrm{photo}} + \mathcal{L}_{\mathrm{smooth}}$}
\newcommand{\expPseudoGTPhotoBuildings}{$\mathcal{L}_{\mathrm{disp}} + \mathcal{L}_{\mathrm{photo}}$}
\newcommand{\expPseudoGTPhotoBuildingsSmooth}{$\mathcal{L}_{\mathrm{disp}}+\mathcal{L}_{\mathrm{photo}}+\mathcal{L}_{\mathrm{smooth}}$}
\newcommand{\expDiachronicStereo}{Diachronic Stereo~\cite{masquil2026diachronic}}

\begin{table}[t]
\centering
\caption{Altitude MAE (in meters) across test sets. Avg. is computed across sets. (a)-(d) are trained on the full SeasonStereo dataset: (a) uses LiDAR-based supervision, and (b)-(d) show incremental SeasonStereo loss configurations; (d) is our final model~\eqref{eq:seasonstereo_loss}.}

\vspace{-0.5em}

\label{table:results_evaluation}
\begin{adjustbox}{max width=\linewidth}
\begin{tabular}{lccccc}
\toprule
\textbf{Experiment} & {\small \textbf{Jacksonville}} & {\small \textbf{Buenos Aires}} & {\small \textbf{Omaha Synch.}} & {\small \textbf{Omaha Diach.}} & {\small \textbf{Avg. \!MAE}} \\ 
\midrule
MonSter~\cite{monster}           & 1.70$\pm$0.45 & 2.32$\pm$0.43 & 0.87$\pm$0.30 & 1.59$\pm$0.67 & 1.62 \\
MonSter++~\cite{monster++}       & 1.53$\pm$0.57 & 2.38$\pm$0.56 & 0.87$\pm$0.41 & 2.93$\pm$3.23 & 1.93 \\
\expDiachronicStereo             & \textbf{1.12$\pm$0.45} & 1.54$\pm$0.21 & 0.75$\pm$0.34 & 0.82$\pm$0.34 & 1.06 \\ 
\midrule
(a) \expGT                           & 1.17$\pm$0.40 & 1.50$\pm$0.18 & \textbf{0.74$\pm$0.31} & 0.80$\pm$0.37 & \textbf{1.05} \\
(b) \expPseudoGT                     & 1.20$\pm$0.43 & \textbf{1.48$\pm$0.16} & 0.76$\pm$0.32 & \textbf{0.79$\pm$0.35} & 1.06 \\
(c) \expPseudoGTPhotoBuildings       & 1.18$\pm$0.40 & 1.58$\pm$0.13 & 0.77$\pm$0.32 & 0.84$\pm$0.38 & 1.09 \\
(d) \expPseudoGTPhotoBuildingsSmooth & 1.15$\pm$0.39 & 1.52$\pm$0.14 & 0.75$\pm$0.32 & \textbf{0.79$\pm$0.36} & \textbf{1.05} \\
\bottomrule
\end{tabular}
\end{adjustbox}
\end{table}

\subsection{Evaluation}

Table~\ref{table:results_evaluation} reports quantitative evaluation results. We compare the zero-shot MonSter~\cite{monster} and MonSter++~\cite{monster++} models with Diachronic Stereo~\cite{masquil2026diachronic} and our SeasonStereo variants (a)-(d), which range from LiDAR-derived disparity supervision to the complete multi-term loss in Eq.~\eqref{eq:seasonstereo_loss}. The complete SeasonStereo loss in Eq.~\eqref{eq:seasonstereo_loss} matches the lowest average MAE, tying with the LiDAR-supervised variant $\mathcal{L}_{\mathrm{LiDAR}}$. This suggests that zero-shot disparity predictions from foundation stereo models can serve as effective pseudo-ground truth, providing a scalable alternative to LiDAR-derived supervision.

Although $\mathcal{L}_{\mathrm{disp}}$ and the complete SeasonStereo formulation (d) obtain similar quantitative results in Table~\ref{table:results_evaluation} on the Omaha Diachronic split, Fig.~\ref{fig:disp_detail} reveals qualitative differences. Using $\mathcal{L}_{\mathrm{disp}}$ alone preserves global scene geometry, but produces smoother object boundaries. In contrast, the final loss yields sharper building contours that are better aligned with the image evidence while preserving global smoothness. This indicates that the photometric and smoothness terms complement the disparity supervision by improving structural sharpness around permanent structures.

\newcommand{\markx}{312}        
\newcommand{\marky}{104}        
\newcommand{\imgw}{363}        
\newcommand{\imgh}{643}        
\newcommand{\markcolor}{magenta}
\newcommand{\marksize}{3pt}
\newcommand{\marklinewidth}{1.2pt}
\newcommand{\markedimage}[1]{%
\begin{tikzpicture}[inner sep=0, outer sep=0]
    \node[inner sep=0, anchor=north west] (img) at (0,0)
        {\includegraphics[width=\linewidth]{#1}};

    \pgfmathsetmacro{\xf}{\markx/\imgw}
    \pgfmathsetmacro{\yf}{\marky/\imgh}

    \coordinate (pTop) at ($(img.north west)!\xf!(img.north east)$);
    \coordinate (pBottom) at (pTop |- img.south west);
    \coordinate (p) at ($(pTop)!\yf!(pBottom)$);

    \draw[\markcolor, line width=\marklinewidth]
        ($(p)+(-\marksize,0)$) -- ($(p)+(\marksize,0)$);
    \draw[\markcolor, line width=\marklinewidth]
        ($(p)+(0,-\marksize)$) -- ($(p)+(0,\marksize)$);
\end{tikzpicture}%
}
\begin{figure}[t]
    \centering
    \setlength{\tabcolsep}{1pt}
    \begin{adjustbox}{max width=\linewidth}
    \begin{tabular}{
        @{}
        *{8}{>{\centering\arraybackslash\scriptsize}m{0.12\linewidth}}
        @{}
    }
        \makecell{Left\\image} &
        \makecell{LiDAR\\disparity} &
        \makecell{MonSter++\\\cite{monster++}} &
        \makecell{Diachronic\\Stereo~\cite{masquil2026diachronic}} &
        \makecell{$\mathcal{L}_{\mathrm{LiDAR}}$\\(a)} &
        \makecell{$\mathcal{L}_{\mathrm{disp}}$\\(b)} &
        \makecell{\!\!\!\!\!$\mathcal{L}_{\mathrm{disp}}+\mathcal{L}_{\mathrm{photo}}$\\(c)} &
        \makecell{Ours, full\\(d)}
        \\[3pt]

        \markedimage{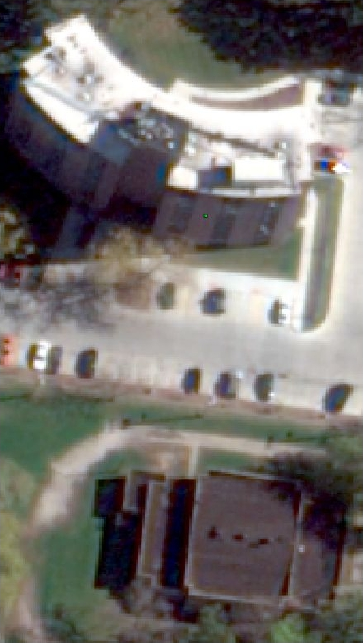} &
        \markedimage{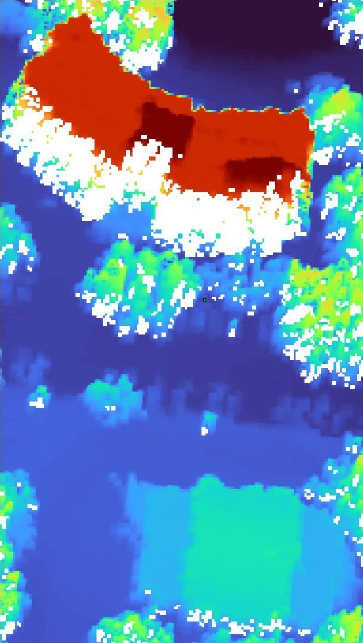} &
        \markedimage{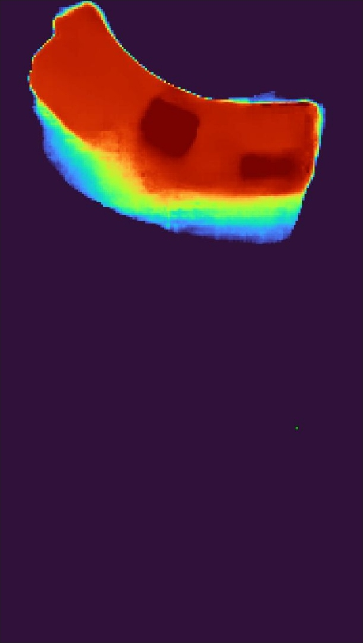} &
        \markedimage{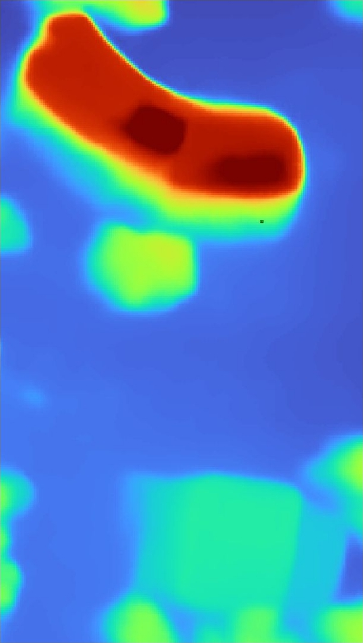} &
        \markedimage{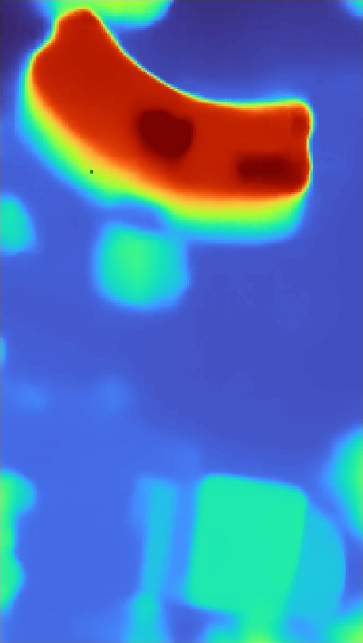} &
        \markedimage{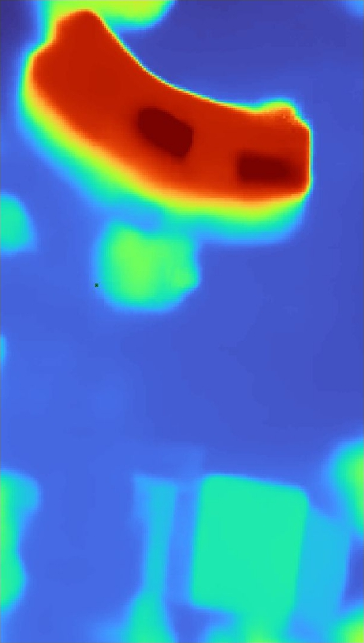} &
        \markedimage{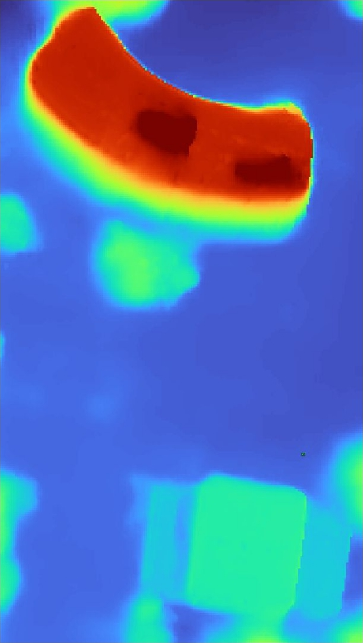} &
        \markedimage{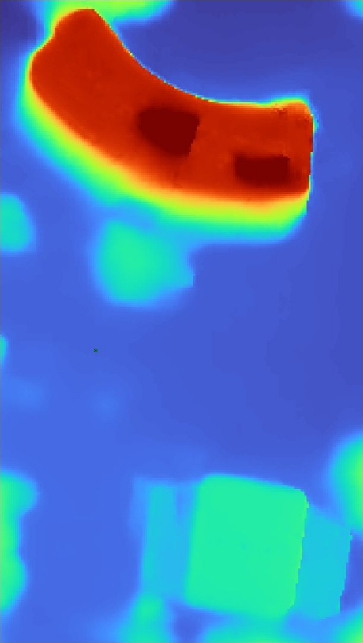}
    \end{tabular}
    \end{adjustbox}
    
    \vspace{-0.5em}
    
    \caption{Qualitative detail from a diachronic test pair in OMA 084, DFC2019~\cite{bosch2019}. MonSter++ fails under diachronic appearance changes, while Diachronic Stereo oversmooths building contours. (a)-(d) are trained on the full SeasonStereo dataset: (a) uses LiDAR-based supervision, and (b)-(d) show incremental SeasonStereo loss configurations; the full model based on Eq.~\eqref{eq:seasonstereo_loss} yields sharper building edges better aligned with image evidence. Magenta markers indicate the building-corner image location.}
    \label{fig:disp_detail}
\end{figure}

\begin{figure}[t]
    \centering
    \includegraphics[width=1\linewidth]{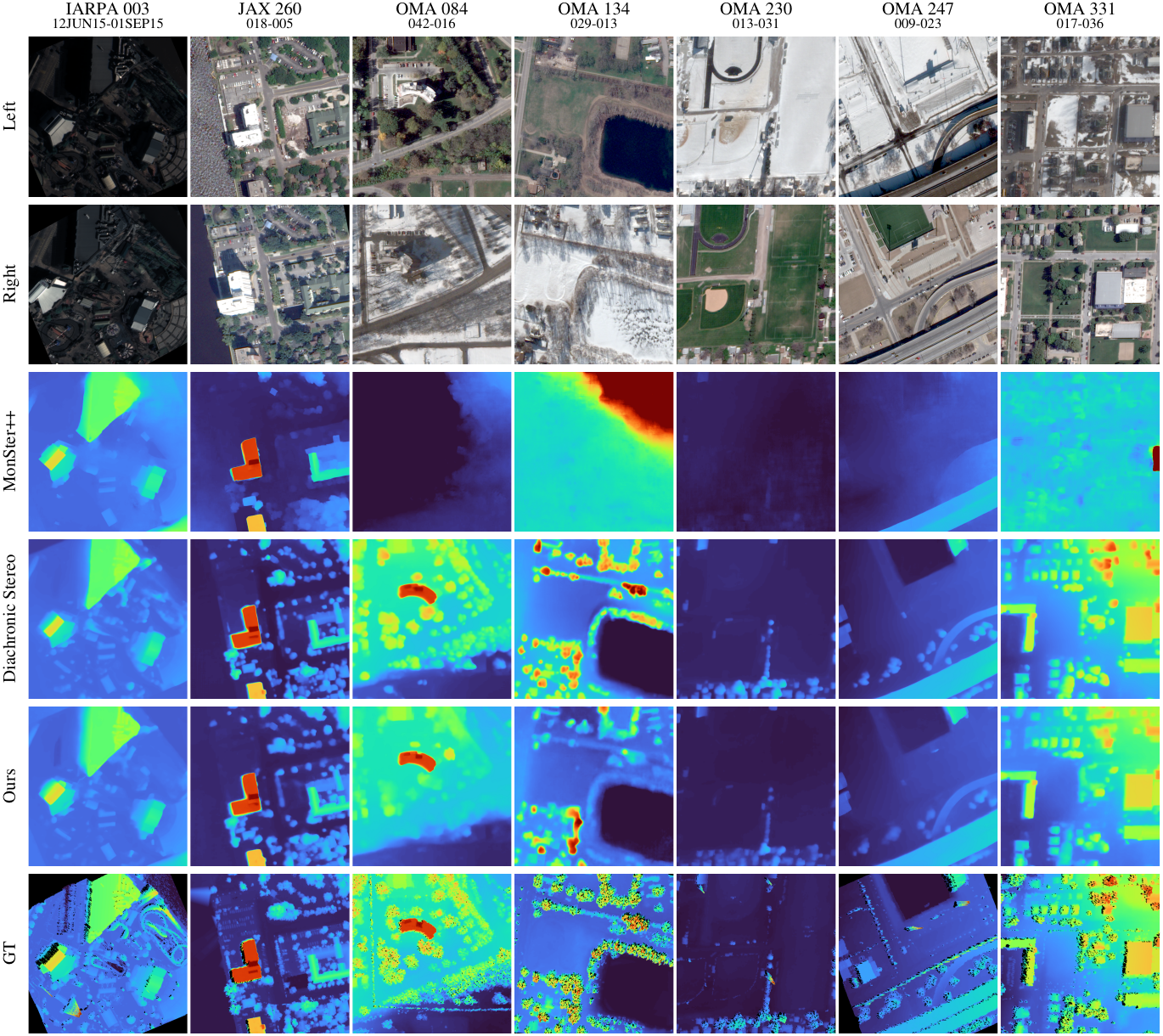}
    
    \vspace{-0.5em}
    
    \caption{Qualitative results of disparity predictions on a selection of hard diachronic image pairs from the test sets, originally listed in Masquil et al.~\cite{masquil2026diachronic}.}
    \label{fig:disparity_predictions}
\end{figure}

\begin{figure}[t]
    \centering
    \includegraphics[width=1\linewidth]{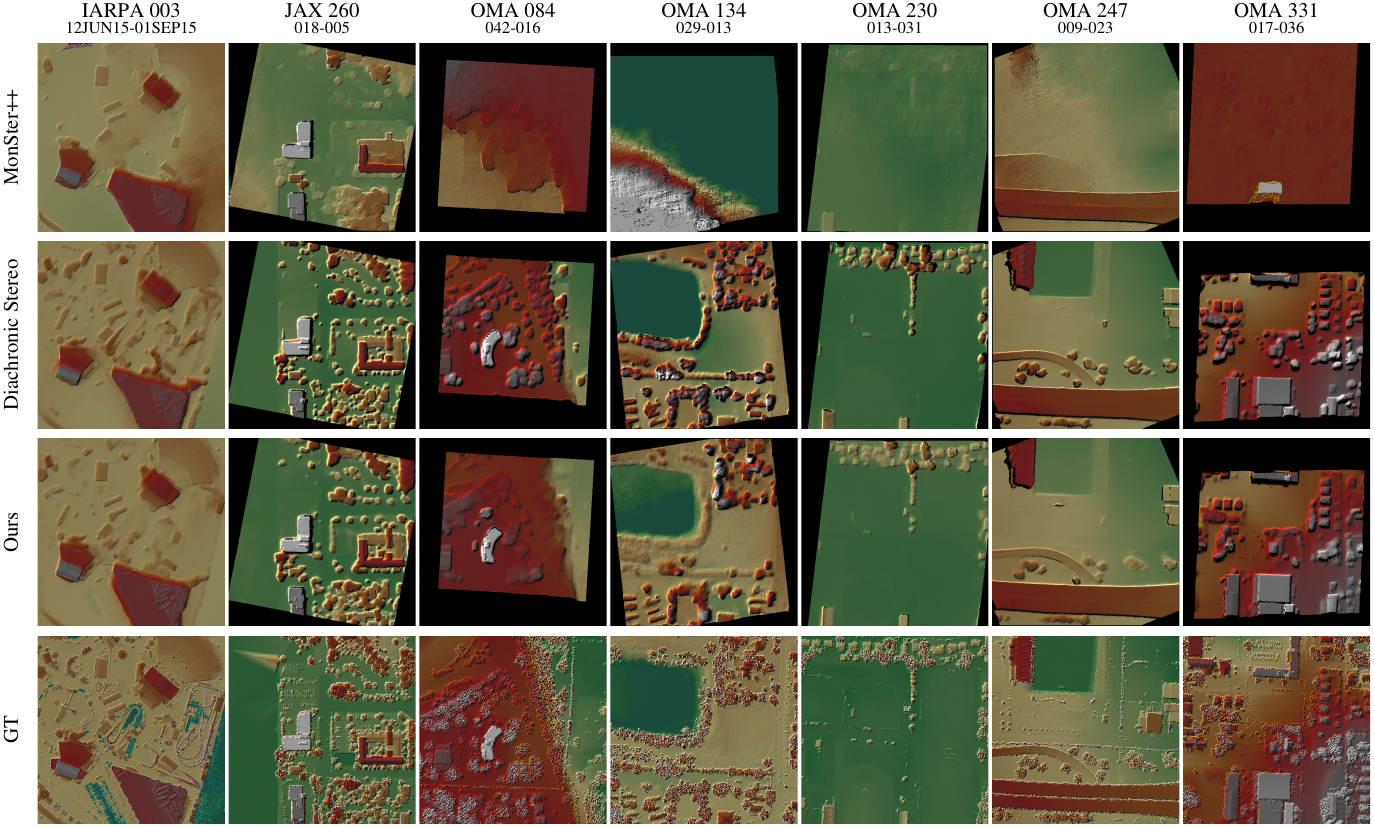}
    
    \vspace{-0.5em}

    \caption{Top to bottom: DSMs reconstructed from disparities predicted by zero-shot MonSter++~\cite{monster++}, Diachronic Stereo~\cite{masquil2026diachronic}, and SeasonStereo (ours), followed by the LiDAR-derived ground-truth DSM. Results are shown for hard diachronic test pairs originally listed by Masquil et al.~\cite{masquil2026diachronic}. Missing values are shown in black.}
    \label{fig:DSM_predictions}
\end{figure}

Figures~\ref{fig:disparity_predictions} and~\ref{fig:DSM_predictions} show further qualitative results in disparity and DSM space. Compared with zero-shot MonSter++ and Diachronic Stereo, SeasonStereo preserves scene geometry under strong seasonal appearance changes while producing sharper building structures and better localized height discontinuities.

\begin{table}[t]
\centering
\caption{Ablation study. We train our full model using \eqref{eq:seasonstereo_loss} on the data subsets described in Sec.~\ref{subsec:data_eval_metrics} to assess the contribution of diachronic synthetic imagery to training.}

\vspace{-0.5em}
    
\label{table:ablation_studies}
\begin{adjustbox}{max width=\linewidth}
\setlength{\tabcolsep}{2pt}
\begin{tabular}{@{}p{2.7cm}cccccc@{}}
\toprule
\textbf{Training data} & \textbf{Number of pairs} & {\small \textbf{Jacksonville}} & {\small \textbf{Buenos Aires}} & {\small \textbf{Omaha Synch.}} & {\small \textbf{Omaha Diach.}} & {\small \textbf{Avg. \!MAE}}  \\ 
\midrule
S1: real only
& $871 \, \text{real}$  & 1.33$\pm$0.48 & 1.63$\pm$0.13 & \textbf{0.75$\pm$0.31} & 1.04$\pm$0.38 & 1.18 \\
S2: +synch.~synth.
& $871 \, \text{real} + 3483 \, \text{synth.}$ & 1.21$\pm$0.39 & \textbf{1.54$\pm$0.19} & 0.78$\pm$0.34 & 0.88$\pm$0.37 & 1.10 \\
S3: +mixed~synth.
& $871 \, \text{real} + 3483 \, \text{synth.}$ & \textbf{1.18$\pm$0.41} & 1.57$\pm$0.16 & \textbf{0.75$\pm$0.31} & \textbf{0.81$\pm$0.35} & \textbf{1.07} \\
\bottomrule
\end{tabular}
\end{adjustbox}
\end{table}

\subsection{Ablation Study}

Table~\ref{table:ablation_studies} evaluates the impact of training-data composition using the complete SeasonStereo loss in Eq.~\eqref{eq:seasonstereo_loss} and the S1/S2/S3 subsets presented in Sec.~\ref{subsec:data_eval_metrics}.
The results show that synthetic augmentation consistently improves performance over training on real synchronic pairs alone, reducing the average MAE from 1.18 in S1 to 1.10 in S2 and 1.07 in S3. The strongest gain appears on the Omaha Diachronic test set, where S3 reduces the MAE from 1.04 to 0.81, indicating that synthetic diachronic training data increases robustness to strong appearance changes. In contrast, the improvement is smaller or non-monotonic on easier settings such as Omaha Synchronic and Buenos Aires, suggesting that the pretrained geometric priors already handle moderate appearance variation reasonably well.

Notably, S3 uses substantially fewer training pairs than the full SeasonStereo training set, yet remains close to the best result reported in Table~\ref{table:results_evaluation}, suggesting that the proposed supervision remains effective with substantially less data. 
\section{Conclusion}
\label{sec:conclusion}

We introduced SeasonStereo, a scalable framework for robust dense stereo matching of multi-date satellite imagery, designed to produce accurate disparity estimates even when input views exhibit strong appearance differences. By generating geometry-preserving seasonal variants from reliable synchronic stereo pairs, SeasonStereo enables diachronic training without requiring aligned real multi-date products or LiDAR-derived dense supervision. The framework combines geometric priors from foundation stereo models with image-consistency cues computed on the corresponding synchronic views, providing an effective alternative to costly ground-truth acquisition.

Our experiments show that SeasonStereo achieves performance on par with LiDAR-supervised diachronic stereo approaches, while recovering sharper fine-scale geometry, particularly along building contours. These results suggest that generative appearance synthesis, when coupled with carefully designed geometric supervision, offers a practical path toward robust large-scale 3D reconstruction from heterogeneous satellite imagery.

Future work will focus on improving the diversity and geometric fidelity of the generated satellite images used in SeasonStereo. Simulating geometry-consistent temporal variations across a wider range of climatic conditions could further increase the reliability and broaden the applicability of synthetic data for dense stereo matching in the satellite domain.



%
%
\bibliographystyle{splncs04}
\bibliography{main}
\end{document}